\documentclass[11pt]{article}

\usepackage[final]{acl}

\usepackage{times}
\usepackage{latexsym}

\usepackage[T1]{fontenc}

\usepackage[utf8]{inputenc}

\usepackage{microtype}

\usepackage{inconsolata}

\usepackage{graphicx}
\usepackage{amsmath}
\usepackage{amssymb}
\usepackage{booktabs}
\usepackage{placeins}
\usepackage{array}
\usepackage[table]{xcolor}
\definecolor{sggray}{gray}{0.92}
\usepackage{stfloats}
\usepackage{multirow}
\usepackage{tabularx}
\usepackage{makecell}
\usepackage[most]{tcolorbox}
\newcommand{\inc}[1]{\textcolor{red}{$\uparrow$#1}}
\newcommand{\dec}[1]{\textcolor{green!60!black}{$\downarrow$#1}}
\newcolumntype{Y}{>{\centering\arraybackslash}m{0.32\textwidth}}
\title{SafeGene: Reusable Adapters for Transferable Safety Alignment}

\author{
  \textbf{Yanghan Wang\textsuperscript{1}},
  \textbf{Zhiqiang Kou\textsuperscript{2}},
  \textbf{Fu Feng\textsuperscript{1}},
  \textbf{Jing Wang\textsuperscript{1*}},
  \textbf{Xin Geng\textsuperscript{1*}} \\
  \textsuperscript{1}Southeast University,
  \textsuperscript{2}The Hong Kong Polytechnic University \\
  \texttt{\{yanghanwang,fufeng,wangjing91,xgeng\}@seu.edu.cn} \\
}

\begin{document}
\maketitle

\begingroup
\renewcommand{\thefootnote}{\fnsymbol{footnote}}
\footnotetext[1]{Corresponding authors.}
\endgroup
\begin{abstract}
Open-weight LLMs are increasingly fine-tuned into customized assistants, but downstream fine-tuning can weaken safety alignment and make models more vulnerable to malicious prompts, even when the training data is not intentionally harmful. This creates a recurring safety recovery problem as target models are repeatedly updated with new task data or user interactions. We propose SafeGene, a reusable safety-adapter module designed for cross-task reuse within each architecture-compatible model family. Rather than treating safety recovery as a model-specific repair step, SafeGene treats safety capability as an independent, reusable adapter representation decoupled from task-specific updates. This representation is obtained from aligned--degraded model discrepancies, refined into task-transferable safety vectors through data-aware layer selection, and expressed in each downstream task-adapted model via few-shot layer-wise coefficient recalibration. Experiments across multiple model families, downstream tasks, and safety judges show that SafeGene-enhanced models reduce harmful response rates while maintaining downstream performance, outperforming representative safe adaptation methods in safety--utility trade-off.
\end{abstract}

\section{Introduction}

\begin{figure}[!t]
\centering
\includegraphics[width=1\linewidth]{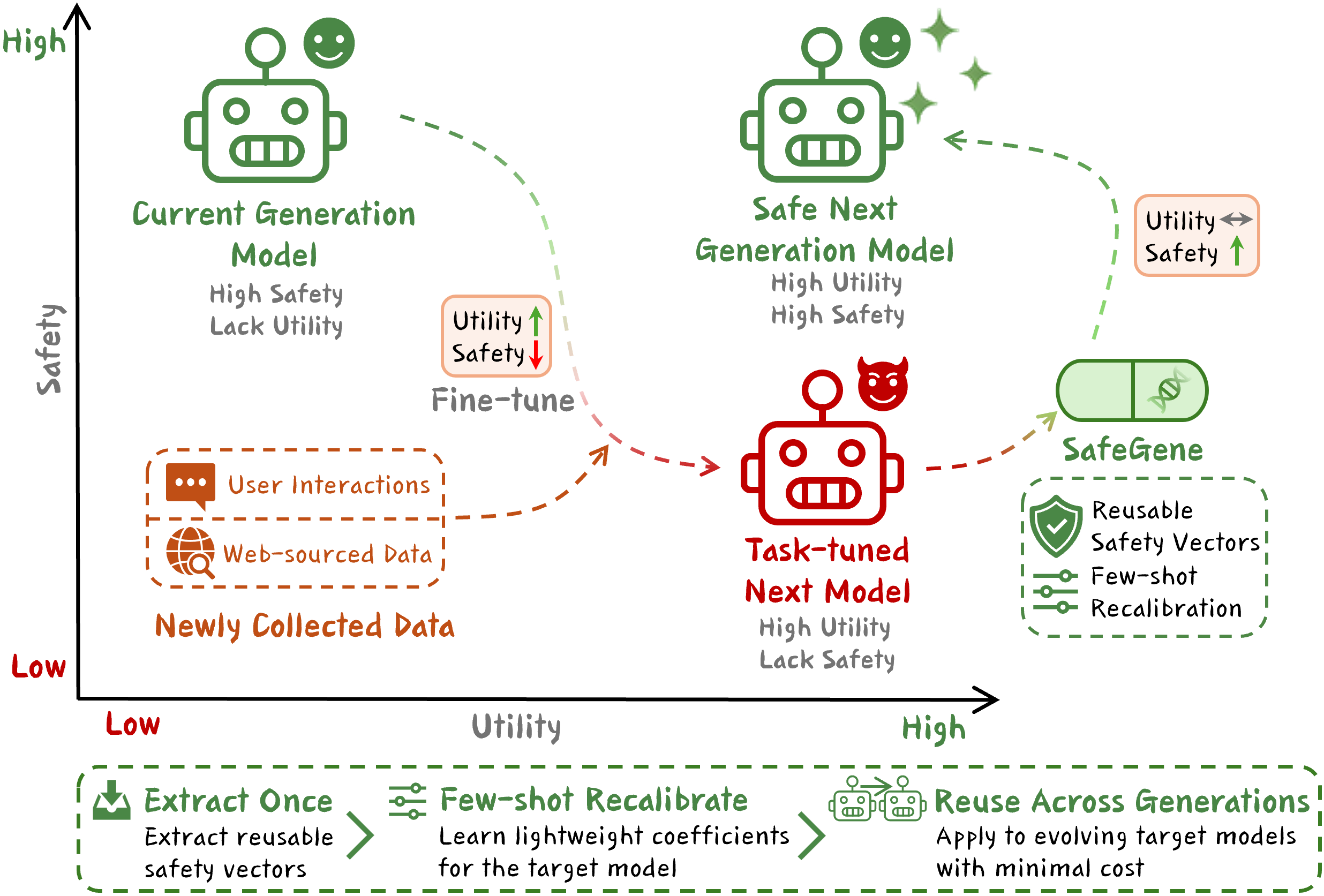}
\caption{
Motivation of SafeGene. Continual fine-tuning on newly collected data improves the utility of next-generation models but can weaken safety alignment. Within each architecture-compatible model family, a SafeGene module is instantiated once from reusable safety vectors and reused across evolving downstream task-adapted models through few-shot recalibration, restoring safety while preserving task utility.
}
\label{fig:mpic}
\end{figure}

Large language models (LLMs) are increasingly customized for high-value domain assistants, such as medical, legal, financial, and enterprise applications. In many practical settings~\citep{hu2022lora,dettmers2023qlora}, developers start from publicly released open-weight models and fine-tune them on private or task-specific datasets to obtain customized target models.

While such customization improves downstream utility, it also introduces a critical safety risk: \textbf{\emph{alignment-acquired safety may not survive downstream adaptation}}. As shown in Figure~\ref{fig:mpic}, task fine-tuning can improve utility while weakening safety. Prior studies~\citep{qi2024fine,fraser2025fine} show that even benign fine-tuning may degrade safety alignment, making fine-tuned models more likely to comply with malicious prompts~\citep{hsu2024safe,huang2024antidote}. This makes post-fine-tuning safety recovery a practical necessity.

A straightforward solution~\citep{ouyang2022training} is to re-align each fine-tuned model with additional safety data. However, this model-specific repair is costly and difficult to scale. In fine-tuning-as-a-service or personalized deployment settings~\citep{hu2022lora,dettmers2023qlora}, model providers and downstream developers often face continuously changing user needs and evolving data sources. These changes, including newly released datasets or newly collected user interactions, may trigger another round of fine-tuning and shift a customized model to a new distribution. \textbf{\emph{Repeating safety alignment or repair for every updated checkpoint introduces extra computational and annotation overhead and may interfere with the task-specific behavior learned during fine-tuning}}~\citep{farn2024safeguard,pan2025survey,thakkar2025combining}. Therefore, the key challenge in practice is not merely how to repair a single unsafe model, but how to repeatedly and efficiently recover safety across changing downstream target models over time.

Existing safety adaptation methods mitigate post-fine-tuning safety degradation at different intervention stages, including alignment-stage defenses~\citep{huang2024vaccine}, fine-tuning-stage modifications~\citep{wang2024backdooralign,huang2024lisa}, and post-hoc repair methods~\citep{huang2024antidote}. However, many of them still rely on model-specific projection~\citep{hsu2024safe}, merging~\citep{djuhera2025safemerge}, or safety-aware re-training procedures~\citep{li2025salora}, making them less suited in practice to repeatedly updated target models where safety recovery should remain reusable, lightweight, and rapidly adapted.

Motivated by this, we ask whether safety alignment can be extracted as a reusable module rather than re-learned after every downstream update. As shown in Figure~\ref{fig:mpic}, within each architecture-compatible model family, a SafeGene module is obtained once by distilling safety capability from the discrepancy between a safely aligned model and a safety-degraded model, and is reused across evolving downstream tasks. Adaptation then only requires lightweight few-shot recalibration of selected safety components, without directly overwriting the target model's task update.

In this paper, we propose \textbf{SafeGene}, a reusable safety-adapter module for cross-task safety transfer within an architecture-compatible model family. SafeGene is built on the principle of decoupling safety recovery from task adaptation: safety capability is represented as an independent adapter module that can be reused across downstream tasks while preserving task-specific updates. The SafeGene module is obtained by distilling safety-relevant adapter updates from the discrepancy between a safely aligned teacher model and a safety-degraded student model, and is refined through data-aware layer selection to retain task-transferable safety vectors. For each downstream task-adapted model, only one scalar coefficient per selected layer is recalibrated using few-shot target-domain data, allowing the reusable safety adapter to be lightly adjusted to the current target distribution without overwriting the task update.

We evaluate SafeGene on five model families, including Qwen2.5-7B and Qwen2.5-1.5B~\citep{qwen2025qwen25technicalreport}, Qwen3-1.7B~\citep{yang2025qwen3}, GLM-4-9B~\citep{glm2024chatglm}, and Llama-3-8B~\citep{grattafiori2024llama}, across four downstream tasks and three safety benchmarks~\citep{zhang2015character,wang2018glue,clark2019boolq,wang2019superglue,ji2023beavertails,zou2023universal,huang2025safety}. Downstream fine-tuning increases average ASR by 7.70\%, while SafeGene reduces average ASR by 11.48\% compared with the fine-tuned target models. At the same time, downstream accuracy remains nearly unchanged after applying SafeGene. SafeGene also outperforms representative safe adaptation baselines, including SafeLoRA~\citep{hsu2024safe}, SafeMERGE~\citep{djuhera2025safemerge}, and SaLoRA~\citep{li2025salora}, achieving the best safety--utility trade-off on Qwen2.5-7B.

Our contributions are summarized as follows:
\begin{itemize}
    \item We formulate post-fine-tuning safety recovery as a reusable safety-transfer problem, targeting deployment settings where customized models may be repeatedly updated with new task data or user interaction data.
    \item We introduce \textbf{SafeGene}, a reusable safety adapter for within-family cross-task safety transfer, obtained from aligned--degraded model discrepancies and adapted to downstream task-adapted models through lightweight coefficient recalibration.
    \item We show that SafeGene-enhanced models effectively mitigate recurring safety degradation after downstream fine-tuning, substantially reducing ASR while preserving downstream accuracy and outperforming representative safe adaptation methods.
\end{itemize}

\FloatBarrier

\section{Related Work}

\noindent\textbf{Safety degradation.} Recent studies have increasingly shown that safety alignment can be fragile under model customization. \citet{fraser2025fine} show that fine-tuning can lower safety and destabilize safety evaluation. \citet{qi2024fine,wang2024backdooralign,huang2024antidote} show that fine-tuning aligned language models can compromise safety even without malicious intent, with both harmful and benign instruction-tuning data weakening safety behavior. \citet{gong2025safety} suggest that LLMs may fail safety alignment under practical interventions such as fine-tuning and model editing. These works show that fine-tuning for downstream tasks can introduce systematic safety risks to both models and users, motivating our study of safety recovery for customized target models.

\par\noindent\textbf{Safe adaptation.} Recent methods mitigate safety degradation at different stages of model adaptation: alignment-stage defenses, such as Vaccine~\citep{huang2024vaccine} and Booster~\citep{huang2025booster}, improve robustness before user fine-tuning; fine-tuning-stage methods, such as BackdoorAlign~\citep{wang2024backdooralign}, Lisa~\citep{huang2024lisa}, and SaLoRA~\citep{li2025salora}, modify the adaptation process itself; and post-hoc methods, such as Antidote~\citep{huang2024antidote} and SafeMERGE~\citep{djuhera2025safemerge}, repair already fine-tuned checkpoints, while Safe LoRA~\citep{hsu2024safe} constrains downstream LoRA updates through projection. These methods provide useful defenses, but projection-based correction may affect task-relevant components when safety and task directions are entangled, merging-based repair often relies on compatible module structures and fixed correction strengths, and safety-aware fine-tuning is less flexible for already fine-tuned or continuously updated target models. This motivates a reusable, lightweight, and rapidly adaptable safety adapter for transferable safety alignment.


\section{Method}

\subsection{Overview}

\begin{figure*}[t]
  \centering
  \includegraphics[width=\textwidth]{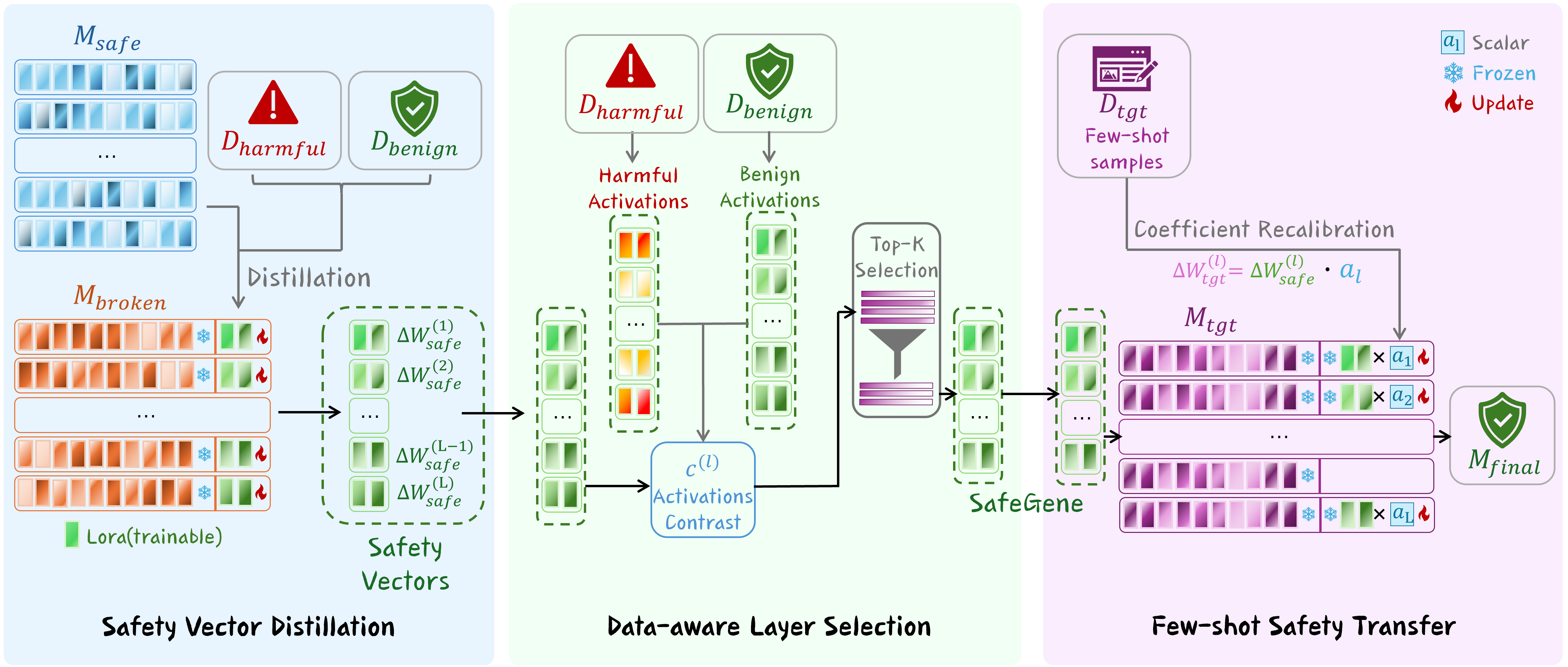}
    \caption{
    Overview of SafeGene. SafeGene decouples safety capability from task-specific updates by representing safety as a reusable module. The module is obtained by distilling safety-relevant vectors from aligned--degraded discrepancies (Section~\ref{sec:safety-vector-distillation}), refined through data-aware layer selection to retain task-transferable components (Section~\ref{sec:data-aware-layer-selection}), and expressed in downstream task-adapted models via few-shot coefficient recalibration (Section~\ref{sec:few-shot-safety-transfer}).
    }
  \label{fig:lpic}
\end{figure*}

We propose \textbf{SafeGene}, a reusable safety-adapter module for cross-task safety transfer within an architecture-compatible model family. Figure~\ref{fig:lpic} provides an overview of  SafeGene pipeline. Reusable safety recovery requires safety capability to be distillable, selectable, and lightly adaptable. SafeGene meets these requirements by distilling safety-relevant adapter updates from aligned--degraded model discrepancies (Section~\ref{sec:safety-vector-distillation}), retaining task-transferable safety vectors through data-aware layer selection (Section~\ref{sec:data-aware-layer-selection}), and recalibrating only layer-wise scalar coefficients for each downstream task-adapted model (Section~\ref{sec:few-shot-safety-transfer}). This enables safety recovery under new target distributions without overwriting task-specific updates.

\subsection{Notations}

Let $M_{\text{safe}}$ denote a safely aligned teacher model, and let $M_{\text{broken}}$ denote a safety-degraded student model with the same architecture, obtained by lightly degrading the safety behavior of $M_{\text{safe}}$. The student model is used only to expose the safety discrepancy from which reusable safety information is distilled, and is independent of downstream target models. We use $\mathcal{D}_h$ and $\mathcal{D}_b$ to denote the source harmful and benign data used during safety distillation. For downstream adaptation, let $M_{\text{tgt}}$ denote a target model with a compatible architecture that has already been fine-tuned on a task-specific dataset.

\subsection{Safety Vector Distillation}
\label{sec:safety-vector-distillation}

The first stage distills safety behavior from the aligned teacher model $M_{\mathrm{safe}}$ into a parameter-efficient adapter attached to the student model $M_{\mathrm{broken}}$. We initialize the candidate layer set $\mathcal{T}_{\mathrm{cand}}$ as the full set of Transformer layers in $M_{\mathrm{broken}}$, and insert a LoRA module~\citep{hu2022lora} into each layer while keeping the backbone parameters frozen. For each candidate layer $\ell$, the induced safety update is represented as $\Delta\mathbf{W}_{\mathrm{safe}}^{(\ell)} = \mathbf{B}^{(\ell)}\mathbf{A}^{(\ell)}$, where $\mathbf{A}^{(\ell)}$ and $\mathbf{B}^{(\ell)}$ are the learned low-rank LoRA matrices. The collection of these layer-wise updates forms the safety adapter and serves as the initial safety-vector bank.

The distillation objective balances safety imitation and benign utility preservation:
\begin{equation}
\begin{aligned}
\mathcal{L}_{\text{distill}}
&\mathrel{=} \lambda_{\text{ce}}\mathcal{L}_{\text{CE}}^{h}
+ \lambda_{\text{kl}}\mathcal{L}_{\text{KL}}^{h} \\
&\mathrel{+} \lambda_{\text{repr}}\mathcal{L}_{\text{repr}}^{h}
+ \lambda_{\text{benign}}\mathcal{L}_{\text{CE}}^{b}
\end{aligned}
\label{eq:distill-loss}
\end{equation}

Here, $\mathcal{L}_{\mathrm{CE}}^{h}$ supervises the student model with safe responses on harmful inputs, $\mathcal{L}_{\mathrm{KL}}^{h}$ matches the output distribution of $M_{\mathrm{safe}}$, and $\mathcal{L}_{\mathrm{repr}}^{h}$ optionally aligns intermediate representations with the teacher. The benign loss $\mathcal{L}_{\mathrm{CE}}^{b}$ prevents the distilled safety adapter from unnecessarily harming general utility. This stage outputs a compact but expressive dense safety-vector bank $\{\Delta\mathbf{W}_{\mathrm{safe}}^{(\ell)}\}_{\ell \in \mathcal{T}_{\mathrm{cand}}}$.

By leveraging the safety discrepancy between $M_{\mathrm{broken}}$ and $M_{\mathrm{safe}}$, we distill and consolidate the safety capability of $M_{\mathrm{safe}}$ into a structured layer-wise safety-vector bank. This bank characterizes how safety behavior is distributed across layers, from which transferable safety vectors are subsequently selected.

\subsection{Data-aware Layer Selection}
\label{sec:data-aware-layer-selection}

Although the dense safety-vector bank extracted in the previous stage already consolidates the core safety information distilled from $M_{\text{safe}}$, it may still contain redundant updates and distribution-specific noise. From a layer-wise perspective, some layers contribute little to safety transfer, while others may even interfere with benign capabilities~\citep{wang2023learngene}. We therefore further select a compact subset of candidate layers from the dense safety-vector bank such that they are more sensitive to harmful inputs while minimizing potential interference with general benign behavior.

To assess the transferability of each layer, we analyze the layer-wise activation statistics of $M_{\mathrm{broken}}$ equipped with the full safety-vector bank on both benign and harmful data. For an input $x$, let $\mathbf{h}_t^{(\ell)}(x)$ denote the final token representation at position $t$ after the $\ell$-th Transformer block. We define the average activation magnitude of layer $\ell$ as
\begin{equation}
  a^{(\ell)}(x) = \frac{1}{T_x} \sum_{t=1}^{T_x} 
  \left\| \mathbf{h}_t^{(\ell)}(x) \right\|_2
  \label{eq:activation-magnitude}
\end{equation}

Let $\mu_{\mathrm{h}}^{(\ell)}$ and $\mu_{\mathrm{b}}^{(\ell)}$ denote the average 
activation magnitudes of layer $\ell$ on harmful and benign data, respectively:
\begin{equation}
\begin{cases}
  \mu_{\mathrm{h}}^{(\ell)} = \mathbb{E}_{x \sim \mathcal{D}_{\mathrm{h}}}
  \left[ a^{(\ell)}(x) \right], \\[2pt]
  \mu_{\mathrm{b}}^{(\ell)} = \mathbb{E}_{x \sim \mathcal{D}_{\mathrm{b}}}
  \left[ a^{(\ell)}(x) \right].
\end{cases}
\label{eq:activation-means}
\end{equation}

We compute the harmful--benign activation contrast:
\begin{equation}
  c^{(\ell)} = \left[ \mu_{\mathrm{h}}^{(\ell)} 
  - \gamma \mu_{\mathrm{b}}^{(\ell)} \right]_+
  \label{eq:activation-contrast}
\end{equation}

where $[z]_+ = \max(z, 0)$, and $\gamma$ controls the penalty on benign activation. The contrast score favors layers that respond more strongly to harmful inputs while remaining less active on benign inputs. Layers with non-positive contrast are discarded by the truncation, since they do not provide clear evidence of harmful-input selectivity.

We rank candidate layers by their activation contrast $c^{(\ell)}$. Intuitively, safety vectors attached to high-contrast layers are more likely to affect safety-relevant behavior while causing less interference with benign task behavior. We therefore select the top-scoring layers as
\begin{equation}
  \mathcal{T} = \mathrm{TopK}_{\ell \in \mathcal{T}_{\mathrm{cand}}}
  \left(c^{(\ell)}\right),
  \label{eq:topk-selection}
\end{equation}
and retain only the corresponding safety vectors:
\begin{equation}
  \left\{\Delta\mathbf{W}_{\mathrm{safe}}^{(\ell)}\right\}_{\ell \in \mathcal{T}} .
  \label{eq:selected-safety-vectors}
\end{equation}
The selected vectors constitute the final SafeGene module.

This step forms SafeGene, a sparse and transferable safety-adapter module.

\subsection{Few-shot Safety Transfer}
\label{sec:few-shot-safety-transfer}

After the safety vectors have been extracted and selected, we transfer SafeGene to a downstream target model $M_{\text{tgt}}$. The target model is not involved in safety distillation; it only receives the selected safety vectors as a reusable safety adapter during transfer. This separation is central to SafeGene: the safety capability distilled from the teacher model is packaged as a transferable adapter and then attached to the downstream target model, thereby enhancing safety while minimizing interference with downstream task performance.

For each selected layer $\ell \in \mathcal{T}$, we apply the transferred safety update as $\Delta W_{\text{tgt}}^{(\ell)} = \alpha_{\ell} \Delta\mathbf{W}_{\mathrm{safe}}^{(\ell)}$, where $\Delta\mathbf{W}_{\mathrm{safe}}^{(\ell)}$ is frozen and only the scalar coefficient $\alpha_{\ell}$ is trainable. We initialize $\alpha_{\ell}=1$, allowing the target model to start from direct reuse of the selected safety vector and then recalibrate its strength using few-shot target-domain safety data.

This transfer interface is highly parameter-efficient. For rank-$r$ LoRA, adapting one layer would normally require $r(d_{\text{in}} + d_{\text{out}})$ trainable parameters, whereas SafeGene only learns one scalar coefficient $\alpha_{\ell}$ per selected layer to recalibrate the reusable safety adapter.

The few-shot adaptation objective is
\begin{equation}
\begin{aligned}
\mathcal{L}_{\text{few}}
&\mathrel{=} \lambda_{\text{ce}}\mathcal{L}_{\text{CE}}
+ \lambda_{\text{anchor}}
\sum_{\ell \in \mathcal{T}}(\alpha_{\ell}-1)^2 \\
&\mathrel{+} \lambda_{\text{sparse}}
\sum_{\ell \in \mathcal{T}}|\alpha_{\ell}|
+ \lambda_{\text{benign}}\mathcal{L}_{\text{benign}} .
\end{aligned}
\label{eq:fewshot-loss}
\end{equation}

The CE and benign terms guide safety learning on harmful examples while preserving benign behavior. The anchor and sparsity regularizers keep useful safety vectors near their original strength and suppress unnecessary ones, yielding a retain-or-suppress trade-off.

After adaptation, the final safety module is
\begin{equation}
  L_{\text{safety}}^{\text{SG}} = \left\{ \alpha_{\ell} \Delta\mathbf{W}_{\mathrm{safe}}^{(\ell)} \right\}_{\ell \in \mathcal{T}}
  \label{eq:final-safety-module}
\end{equation}

At inference, the final model is composed as
\begin{equation}
  M_{\text{final}} = M_{\text{tgt}} \oplus L_{\text{safety}}^{\text{SG}}
  \label{eq:deployment-model}
\end{equation}

where $\oplus$ denotes adapter composition through additive weight updates. If the downstream target model is represented as a task adapter $L_{\text{task}}$ on top of a backbone model, the composition can equivalently be written as
\begin{equation}
  M_{\text{final}} = M_{\text{backbone}} \oplus L_{\text{task}} \oplus L_{\text{safety}}^{\text{SG}}
  \label{eq:deployment-model-full}
\end{equation}

By attaching SafeGene to the downstream target model, we transfer the safety capability distilled from the teacher model as a reusable safety adapter while preserving the target model's task performance as much as possible. Since only a small number of layer-wise coefficients are learned during transfer, the adaptation remains lightweight and parameter-efficient.

\section{Experiments}
\subsection{Experimental Setup}

\noindent\textbf{Models.}
We evaluate SafeGene on five instruction-tuned LLMs with different model families and scales: Qwen2.5-7B and Qwen2.5-1.5B~\citep{qwen2025qwen25technicalreport}, Qwen3-1.7B~\citep{yang2025qwen3}, GLM-4-9B~\citep{glm2024chatglm}, and Llama-3-8B~\citep{grattafiori2024llama}. For each model, we compare three states: the original aligned model (\textit{Base}), its downstream fine-tuned version (\textit{Fine-tuned}), and the SafeGene-enhanced model (\textit{SG}), which incorporates SafeGene into the fine-tuned model and learns only the layer-wise coefficients.

\par\noindent\textbf{Downstream tasks.}
We consider four representative classification and reasoning tasks: AG News~\citep{zhang2015character}, SST2, MNLI, and BoolQ~\citep{clark2019boolq}. SST2 and MNLI are from GLUE~\citep{wang2018glue}, while BoolQ is from SuperGLUE~\citep{wang2019superglue}. For each model family, SafeGene is extracted once from the aligned--degraded model pair and reused across four downstream fine-tuned checkpoints; only $\alpha$ is recalibrated per target. We use 9,000 training examples for each downstream fine-tuning run and report accuracy as the utility metric.

\par\noindent\textbf{Safety evaluation.}
We evaluate safety on three harmful instruction benchmarks: BeaverTails~\citep{ji2023beavertails}, AdvBench~\citep{zou2023universal}, and DirectRefusal~\citep{huang2025safety}. For each benchmark, we generate model responses and compute attack success rate (ASR), where a lower ASR indicates stronger safety. Unless otherwise specified, we use Beaver-dam-7B, a BeaverTails-trained QA moderation model~\citep{ji2023beavertails}, as the main safety judge. To assess evaluator robustness, we additionally use Qwen3.5-Flash as an API-based judge to evaluate responses generated by Qwen2.5-7B, with 200 sampled examples from each safety benchmark. For this auxiliary evaluation, we report the number of successful attacks.

\begin{table*}[!t]
\centering
\tiny
\setlength{\tabcolsep}{2.2pt}
\renewcommand{\arraystretch}{1.08}
\resizebox{\textwidth}{!}{
\begin{tabular}{@{}ll c >{\columncolor{sggray}}c c c c c c c c >{\columncolor{sggray}}c >{\columncolor{sggray}}c >{\columncolor{sggray}}c l@{}}
\toprule
\multirow{2}{*}{\textbf{Model}}
& \multirow{2}{*}{\textbf{Task}}
& \multicolumn{3}{c}{\textbf{Accuracy}}
& \multicolumn{3}{c}{\textbf{ASR$_{\text{Base}}$ $\downarrow$}}
& \multicolumn{3}{c}{\textbf{ASR$_{\text{Fine-tuned}}$ $\downarrow$}}
& \multicolumn{3}{c}{\textbf{ASR$_{\text{SG}}$ $\downarrow$}}
& \multirow{2}{*}{\textbf{$\Delta\overline{\text{ASR}}$}} \\
\cmidrule(lr){3-5}
\cmidrule(lr){6-8}
\cmidrule(lr){9-11}
\cmidrule(lr){12-14}
& & \textbf{FT} & \textbf{SG} & \textbf{$\Delta$}
& \textbf{BT} & \textbf{Adv} & \textbf{DR}
& \textbf{BT} & \textbf{Adv} & \textbf{DR}
& \textbf{BT} & \textbf{Adv} & \textbf{DR}
& \\
\midrule
\multirow{4}{*}{Qwen2.5-7B}
& AG News & 90.75 & 90.91 & \inc{0.16}
& \multirow{4}{*}{38.08} & \multirow{4}{*}{24.23} & \multirow{4}{*}{32.76}
& 50.61 & 33.27 & 45.39
& 34.28 & 30.77 & 32.63
& \dec{10.53} \\
& SST2 & 95.87 & 96.56 & \inc{0.69}
& & &
& 56.43 & 38.27 & 53.03
& 34.85 & 32.88 & 37.24
& \dec{14.25} \\
& MNLI & 88.52 & 88.61 & \inc{0.09}
& & &
& 43.39 & 27.12 & 38.03
& 35.03 & 27.69 & 30.66
& \dec{5.05} \\
& BoolQ & 87.98 & 87.86 & \dec{0.12}
& & &
& 47.89 & 35.96 & 44.87
& 34.16 & 27.69 & 31.84
& \dec{11.68} \\
\midrule
\multirow{4}{*}{Qwen2.5-1.5B}
& AG News & 89.42 & 88.91 & \dec{0.51}
& \multirow{4}{*}{45.82} & \multirow{4}{*}{36.15} & \multirow{4}{*}{40.92}
& 68.49 & 48.85 & 62.76
& 38.72 & 29.42 & 33.95
& \dec{26.00} \\
& SST2 & 95.53 & 95.30 & \dec{0.23}
& & &
& 55.45 & 43.65 & 52.63
& 32.26 & 21.35 & 29.61
& \dec{22.84} \\
& MNLI & 84.23 & 84.45 & \inc{0.22}
& & &
& 51.93 & 41.54 & 48.82
& 32.49 & 27.31 & 30.00
& \dec{17.50} \\
& BoolQ & 80.28 & 79.63 & \dec{0.65}
& & &
& 52.28 & 39.81 & 45.66
& 31.62 & 19.04 & 27.63
& \dec{19.82} \\
\midrule
\multirow{4}{*}{Qwen3-1.7B}
& AG News & 88.79 & 87.97 & \dec{0.82}
& \multirow{4}{*}{33.99} & \multirow{4}{*}{21.54} & \multirow{4}{*}{30.13}
& 48.24 & 30.00 & 45.26
& 37.10 & 27.69 & 31.71
& \dec{9.00} \\
& SST2 & 95.53 & 95.07 & \dec{0.46}
& & &
& 61.51 & 37.12 & 61.18
& 46.16 & 28.08 & 42.24
& \dec{14.44} \\
& MNLI & 84.30 & 83.90 & \dec{0.40}
& & &
& 41.89 & 26.15 & 36.97
& 36.12 & 24.23 & 31.84
& \dec{4.27} \\
& BoolQ & 82.54 & 82.94 & \inc{0.40}
& & &
& 44.26 & 27.88 & 41.97
& 34.28 & 20.96 & 30.53
& \dec{9.45} \\
\midrule
\multirow{4}{*}{GLM-4-9B}
& AG News & 89.76 & 95.68 & \inc{5.92}
& \multirow{4}{*}{20.77} & \multirow{4}{*}{28.85} & \multirow{4}{*}{19.74}
& 18.64 & 23.46 & 18.42
& 9.35 & 8.08 & 9.61
& \dec{11.16} \\
& SST2 & 95.53 & 94.61 & \dec{0.92}
& & &
& 24.93 & 30.38 & 26.84
& 14.43 & 16.92 & 16.18
& \dec{11.54} \\
& MNLI & 88.74 & 89.47 & \inc{0.73}
& & &
& 24.64 & 23.85 & 24.34
& 10.85 & 7.12 & 10.92
& \dec{14.65} \\
& BoolQ & 99.63 & 99.36 & \dec{0.27}
& & &
& 18.23 & 32.12 & 18.03
& 13.21 & 12.88 & 12.63
& \dec{9.89} \\
\midrule
\multirow{4}{*}{Llama-3-8B}
& AG News & 91.39 & 90.18 & \dec{1.21}
& \multirow{4}{*}{40.74} & \multirow{4}{*}{33.46} & \multirow{4}{*}{35.92}
& 41.49 & 34.62 & 38.68
& 38.32 & 30.00 & 35.53
& \dec{3.65} \\
& SST2 & 95.87 & 95.76 & \dec{0.11}
& & &
& 49.45 & 35.77 & 45.79
& 47.89 & 31.73 & 41.45
& \dec{3.31} \\
& MNLI & 88.04 & 87.67 & \dec{0.37}
& & &
& 46.62 & 45.19 & 42.50
& 42.12 & 33.85 & 35.13
& \dec{7.74} \\
& BoolQ & 88.53 & 87.52 & \dec{1.01}
& & &
& 35.60 & 33.46 & 33.16
& 35.26 & 28.46 & 30.13
& \dec{2.79} \\
\midrule
\multicolumn{15}{c}{\textbf{Average over four downstream tasks}} \\
\midrule
Qwen2.5-7B & Avg. & 90.78 & 90.99 & \inc{0.21}
& 38.08 & 24.23 & 32.76
& 49.58 & 33.66 & 45.33
& 34.58 & 29.76 & 33.09
& \dec{10.38} \\
Qwen2.5-1.5B & Avg. & 87.37 & 87.07 & \dec{0.29}
& 45.82 & 36.15 & 40.92
& 57.04 & 43.46 & 52.47
& 33.77 & 24.28 & 30.30
& \dec{21.54} \\
Qwen3-1.7B & Avg. & 87.79 & 87.47 & \dec{0.32}
& 33.99 & 21.54 & 30.13
& 48.98 & 30.29 & 46.35
& 38.42 & 25.24 & 34.08
& \dec{9.29} \\
GLM-4-9B & Avg. & 93.42 & 94.78 & \inc{1.37}
& 20.77 & 28.85 & 19.74
& 21.61 & 27.45 & 21.91
& 11.96 & 11.25 & 12.34
& \dec{11.81} \\
Llama-3-8B & Avg. & 90.96 & 90.28 & \dec{0.68}
& 40.74 & 33.46 & 35.92
& 43.29 & 37.26 & 40.03
& 40.90 & 31.01 & 35.56
& \dec{4.37} \\
\midrule
\textbf{Overall} & \textbf{Avg.} & \textbf{90.06} & \textbf{90.12} & \textbf{\inc{0.06}}
& \textbf{35.88} & \textbf{28.85} & \textbf{31.89}
& \textbf{44.10} & \textbf{34.42} & \textbf{41.22}
& \textbf{31.93} & \textbf{24.31} & \textbf{29.07}
& \textbf{\dec{11.48}} \\
\bottomrule
\end{tabular}
}
\caption{
Per-task results across five model families. BT, Adv, and DR denote BeaverTails, AdvBench, and DirectRefusal, respectively. $\Delta$ denotes Acc$_{\text{SG}}$ $-$ Acc$_{\text{Fine-tuned}}$, and $\Delta\overline{\text{ASR}}$ denotes ASR$_{\text{SG}}$ $-$ ASR$_{\text{Fine-tuned}}$ averaged over the three safety benchmarks. ASR$_{\text{Base}}$ is task-independent because it is measured before downstream fine-tuning. The bottom block reports averages over four downstream tasks and the overall average across model families.
}
\label{tab:per-task-results-five-models}
\end{table*}

\subsection{Main Results: Safety Transfer after Downstream Fine-tuning}
Table~\ref{tab:per-task-results-five-models} reports the per-task results across five model families, four downstream tasks, and three safety benchmarks. For each model and task, we compare the fine-tuned model with the SafeGene-enhanced model in terms of downstream accuracy and attack success rate (ASR). ASR is reported separately on BeaverTails, AdvBench, and DirectRefusal using Beaver-dam-7B as the judge, while $\Delta\overline{\text{ASR}}$ summarizes the average ASR change over the three safety benchmarks. The bottom block reports four-task averages for each model family and the overall average across all five model families.

The results show a clear safety--utility pattern. Across five model families, downstream fine-tuning often increases ASR compared with the original aligned models, indicating that task adaptation can weaken safety behavior. On average, the fine-tuned target models achieve 90.06\% downstream accuracy, while their average ASR increases from 32.21\% to 39.91\%. After applying SafeGene, average ASR drops to 28.44\%, corresponding to an absolute reduction of 11.48 points relative to the fine-tuned target models. Notably, this safety improvement does not come at the cost of downstream utility: SafeGene achieves an average accuracy of 90.12\%, which is nearly identical to the 90.06\% of the fine-tuned target models.

The per-task results further show that the safety improvement is not merely an artifact of averaging. SafeGene reduces $\Delta\overline{\text{ASR}}$ for every model family and on almost all downstream tasks, while keeping downstream accuracy largely unchanged. For example, on Qwen2.5-7B, SafeGene reduces average ASR on AG News, SST2, MNLI, and BoolQ by 10.53, 14.25, 5.05, and 11.68 points, respectively, with only minor accuracy changes. Similar trends are observed on Qwen2.5-1.5B, Qwen3-1.7B, GLM-4-9B, and Llama-3-8B. These results demonstrate that SafeGene consistently restores safety across model families and downstream tasks through lightweight coefficient recalibration, while preserving task-specific performance.

\begin{figure}[!t]
\centering
\includegraphics[width=\columnwidth]{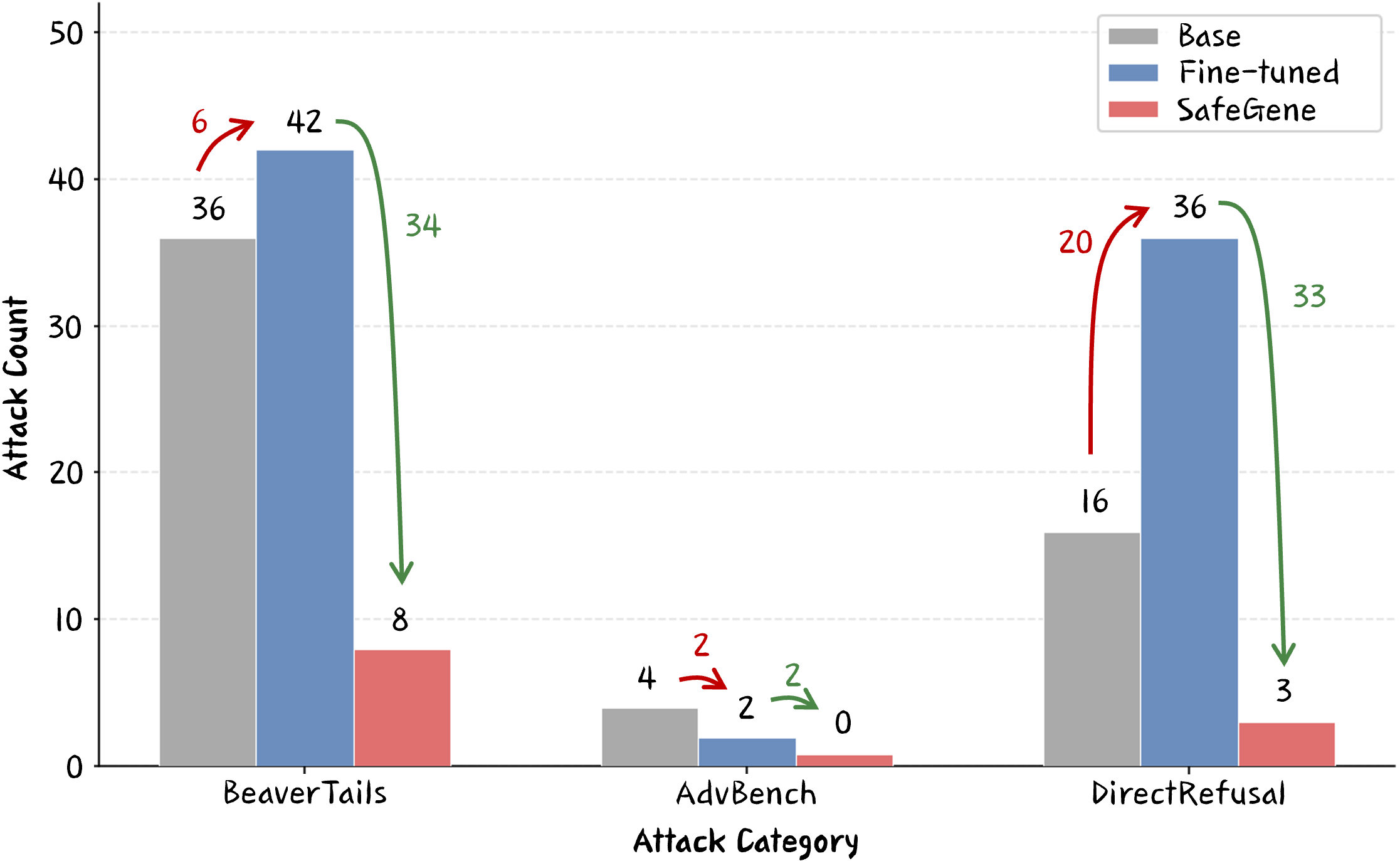}
\caption{
API-based safety evaluation on Qwen2.5-7B using Qwen3.5-Flash as the judge. The figure reports the number of successful attacks on 200 sampled examples from each safety benchmark, aggregated over four downstream tasks. SafeGene substantially reduces successful attacks compared with the fine-tuned model.
}
\label{fig:api-judge}
\end{figure}

We further evaluate Qwen2.5-7B using an API-based safety judge to examine whether the observed trend depends on the choice of evaluator. For this auxiliary evaluation, we sample 200 examples from each of three safety benchmarks for each downstream task, resulting in 2,400 evaluations in total across four tasks. As shown in Figure~\ref{fig:api-judge}, downstream fine-tuning increases the number of successful attacks from 56 to 80, whereas SafeGene reduces it to 11 in this setting. This trend is largely consistent with the Beaver-dam-7B evaluation, further confirming that SafeGene improves safety after downstream fine-tuning.

\subsection{Comparison with Existing Safe Adaptation Methods}
\label{sec:comparison1}

We compare SafeGene with three representative safe adaptation methods on Qwen2.5-7B: SafeLoRA~\citep{hsu2024safe}, SafeMERGE~\citep{djuhera2025safemerge}, and SaLoRA~\citep{li2025salora}. For fairness, all methods use the same fine-tuned checkpoints, generation settings, safety benchmarks, and judge, and all baselines use their default hyperparameters. We report downstream accuracy, BeaverTails ASR, DirectRefusal ASR, and their average ASR, all judged by Beaver-dam-7B.

\begin{table}[!t]
\centering
\small
\setlength{\tabcolsep}{2.0pt}
\renewcommand{\arraystretch}{1.03}
\begin{tabular*}{\columnwidth}{@{\extracolsep{\fill}}l l c c c c@{}}
\toprule
\textbf{Task} 
& \textbf{Method} 
& \textbf{Acc $\uparrow$} 
& \textbf{BT $\downarrow$} 
& \textbf{DR $\downarrow$}
& \textbf{$\overline{\text{ASR}}$ $\downarrow$} \\
\midrule
\multirow{5}{*}{\rotatebox[origin=c]{90}{\textbf{AG News}}}
& Fine-tuned & 90.75 & 50.61 & 45.39 & 48.00 \\
& SafeLoRA   & 90.18 & 44.49 & 37.24 & 40.87 \\
& SafeMERGE  & 89.51 & 41.37 & 36.58 & 38.97 \\
& SaLoRA     & 90.50 & 39.07 & 34.21 & 36.64 \\
& \textbf{SafeGene (Ours)} 
& \textbf{\underline{90.91}} 
& \textbf{\underline{34.28}} 
& \textbf{\underline{32.63}} 
& \textbf{\underline{33.45}} \\
\midrule
\multirow{5}{*}{\rotatebox[origin=c]{90}{\textbf{SST2}}}
& Fine-tuned & 95.87 & 56.43 & 53.03 & 54.73 \\
& SafeLoRA   & \underline{96.67} & 42.41 & 39.87 & 41.14 \\
& SafeMERGE  & 95.76 & 41.03 & \underline{35.00} & 38.02 \\
& SaLoRA     & 96.44 & 38.78 & 35.13 & 36.95 \\
& \textbf{SafeGene (Ours)} 
& \textbf{96.56} 
& \textbf{\underline{34.85}} 
& \textbf{37.24} 
& \textbf{\underline{36.05}} \\
\midrule
\multirow{5}{*}{\rotatebox[origin=c]{90}{\textbf{MNLI}}}
& Fine-tuned & 88.52 & 43.39 & 38.03 & 40.71 \\
& SafeLoRA   & 88.50 & 41.78 & 40.00 & 40.89 \\
& SafeMERGE  & 88.97 & 40.62 & 33.16 & 36.89 \\
& SaLoRA     & \underline{89.61} & 40.68 & 35.79 & 38.23 \\
& \textbf{SafeGene (Ours)} 
& \textbf{88.61} 
& \textbf{\underline{35.03}} 
& \textbf{\underline{30.66}} 
& \textbf{\underline{32.84}} \\
\midrule
\multirow{5}{*}{\rotatebox[origin=c]{90}{\textbf{BoolQ}}}
& Fine-tuned & 87.98 & 47.89 & 44.87 & 46.38 \\
& SafeLoRA   & 86.97 & 42.70 & 36.58 & 39.64 \\
& SafeMERGE  & \underline{89.10} & 39.64 & 34.61 & 37.12 \\
& SaLoRA     & 86.90 & 38.32 & 32.76 & 35.54 \\
& \textbf{SafeGene (Ours)} 
& \textbf{87.86} 
& \textbf{\underline{34.16}} 
& \textbf{\underline{31.84}} 
& \textbf{\underline{33.00}} \\
\midrule
\multirow{5}{*}{\rotatebox[origin=c]{90}{\textbf{Avg.}}}
& Fine-tuned & 90.78 & 49.58 & 45.33 & 47.46 \\
& SafeLoRA   & 90.58 & 42.85 & 38.42 & 40.63 \\
& SafeMERGE  & 90.84 & 40.67 & 34.84 & 37.75 \\
& SaLoRA     & 90.86 & 39.21 & 34.47 & 36.84 \\
& \textbf{SafeGene (Ours)} 
& \textbf{\underline{90.99}} 
& \textbf{\underline{34.58}} 
& \textbf{\underline{33.09}} 
& \textbf{\underline{33.84}} \\
\bottomrule
\end{tabular*}
\caption{
Comparison with existing safe adaptation methods on Qwen2.5-7B. BT and DR denote BeaverTails and DirectRefusal, respectively. $\overline{\text{ASR}}$ is the average ASR over BT and DR. Best results are underlined, and our method is shown in bold.
}
\label{tab:comparison}
\end{table}

Table~\ref{tab:comparison} shows that all safety adaptation methods reduce ASR compared with the fine-tuned model, but SafeGene achieves the strongest safety recovery. On average, SafeLoRA, SafeMERGE, and SaLoRA reduce the two-benchmark average ASR to 40.63\%, 37.75\%, and 36.84\%, respectively, while SafeGene further reduces it to 33.84\%. The improvement is consistent across all four downstream tasks, where SafeGene obtains the lowest average ASR in every task group.

The advantage of SafeGene comes from its explicit decoupling of safety capability from task-specific updates. Rather than constraining, merging, or re-training the task update, SafeGene represents safety as a reusable adapter whose expression strength can be recalibrated while preserving task behavior. This design benefits safety recovery in three complementary ways: aligned--degraded model discrepancies expose clearer safety-relevant updates; data-aware layer selection removes noisy or weakly transferable vectors before reuse; and few-shot coefficient recalibration adjusts the strength of selected safety vectors for each downstream distribution, reducing both under-correction and over-correction. In contrast, SafeLoRA may compress task-relevant update space through projection, SafeMERGE relies on fixed merging strengths that are not explicitly recalibrated for the current task, and SaLoRA is less flexible for already fine-tuned or continuously updated target models. These differences help explain why SafeGene-enhanced models achieve stronger safety recovery while preserving downstream utility across the results in Table~\ref{tab:comparison}.

\subsection{Ablation Study: Layer Selection and Few-shot Safety Transfer}

We conduct two ablation studies on Qwen2.5-7B to examine two key design choices in SafeGene: data-aware layer selection and few-shot safety transfer. For efficiency, all safety results in this subsection are evaluated on BeaverTails. We report averaged results over four downstream tasks for layer selection and per-task results for few-shot transfer.

\begin{figure}[!t]
\centering
\includegraphics[width=1\linewidth]{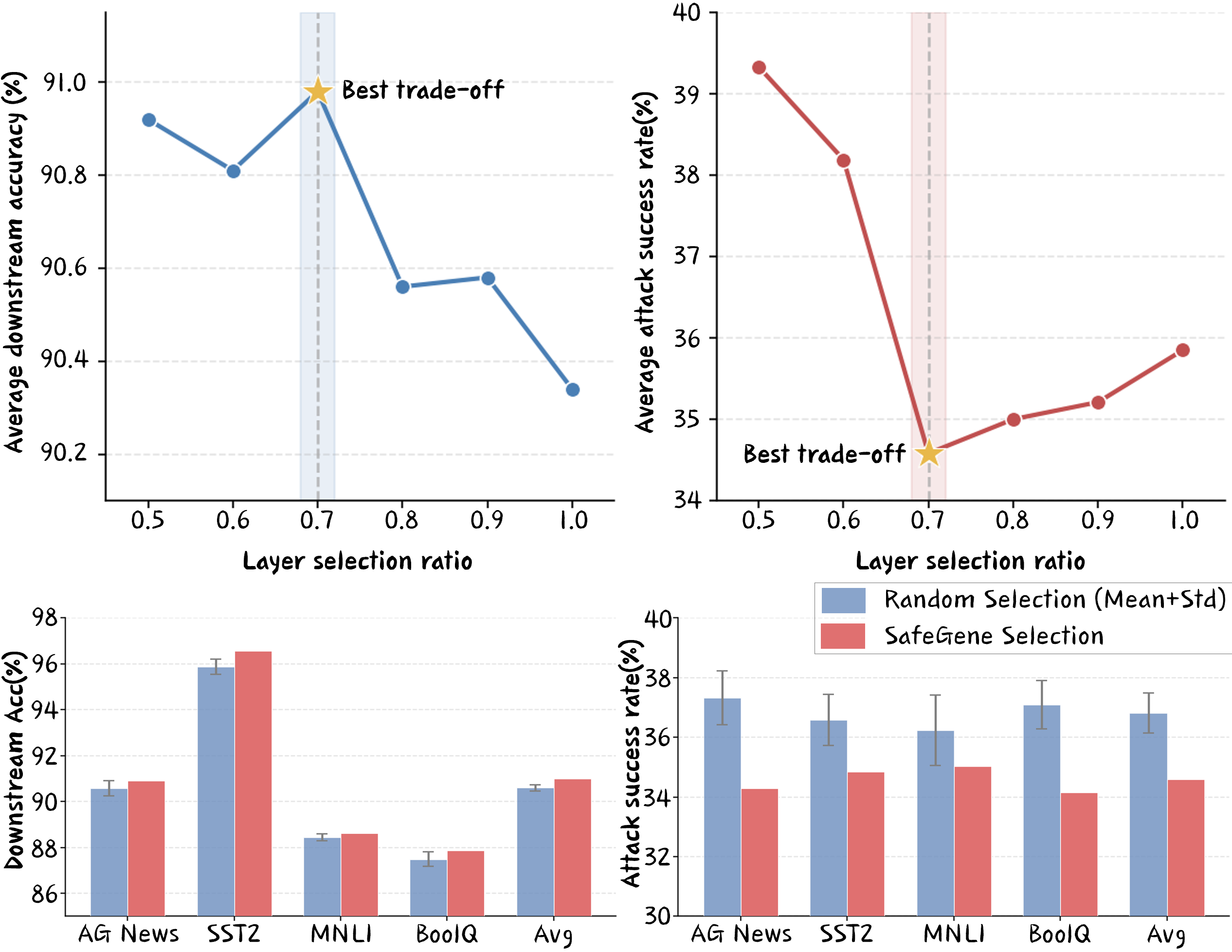}
\caption{
Ablation study of data-aware layer selection on Qwen2.5-7B. 
Top: effect of layer selection ratio on average accuracy and ASR. 
Bottom: comparison with random selection under the same ratio. 
Safety is evaluated on BeaverTails using Beaver-dam-7B.
}
\label{fig:ablation-ratio}
\end{figure}

\paragraph{Effect of layer selection.}
We vary the retained ratio of distilled safety vectors from 1.0 to 0.5 and compare data-aware selection with random selection. Figure~\ref{fig:ablation-ratio} shows a non-monotonic trend: a ratio of 0.7 improves accuracy and lowers ASR, while further pruning hurts safety. At this ratio, data-aware selection outperforms random selection with lower ASR and comparable accuracy. These results show that safety transfer benefits from a compact, data-aware safety-vector set, as data-aware selection can filter noisy vectors and reduce interference with downstream tasks.

\paragraph{Effect of few-shot safety transfer.}
We next examine whether few-shot safety transfer is necessary after layer selection. We compare the full SafeGene pipeline at ratio 0.7 against the results obtained without few-shot safety transfer, where the same selected layers are used but all layer-wise coefficients are fixed to 1. Table~\ref{tab:ablation-fst} shows that the full method achieves higher downstream accuracy on all four tasks and a lower average ASR overall. Few-shot safety transfer better adapts the distilled safety vectors to each downstream model, allowing SafeGene to more flexibly accommodate different dataset distributions and further improve the overall safety--utility trade-off.

\begin{table}[!t]
\centering
\small
\begin{tabular*}{\columnwidth}{@{\extracolsep{\fill}}lcccc@{}}
\toprule
\textbf{Task}
& \multicolumn{2}{c}{\textbf{No FST}}
& \multicolumn{2}{c}{\textbf{SafeGene$^\dagger$}} \\
\cmidrule(lr){2-3} \cmidrule(lr){4-5}
& \textbf{Acc $\uparrow$} & \textbf{ASR $\downarrow$}
& \textbf{Acc $\uparrow$} & \textbf{ASR $\downarrow$} \\
\midrule
AG News 
& 90.36 & 35.26 
& 90.91{\scriptsize(\inc{0.55})} 
& 34.28{\scriptsize(\dec{0.98})} \\

SST2    
& 95.64 & 36.35 
& 96.56{\scriptsize(\inc{0.92})} 
& 34.85{\scriptsize(\dec{1.50})} \\

MNLI    
& 88.55 & 36.47 
& 88.61{\scriptsize(\inc{0.06})} 
& 35.03{\scriptsize(\dec{1.44})} \\

BoolQ   
& 87.22 & 35.72 
& 87.86{\scriptsize(\inc{0.64})} 
& 34.16{\scriptsize(\dec{1.56})} \\
\bottomrule
\end{tabular*}
\caption{
Ablation on few-shot safety transfer. No FST fixes all selected layer-wise coefficients to 1, while SafeGene$^\dagger$ performs coefficient recalibration. Values in parentheses denote changes relative to No FST. ASR is evaluated on BeaverTails using Beaver-dam-7B.
}
\label{tab:ablation-fst}
\end{table}
\section{Conclusion}

In this paper, we studied post-fine-tuning safety recovery in settings where customized LLMs may be repeatedly updated with new task data or user interactions. We proposed \textbf{SafeGene}, a reusable safety-adapter module for within-family cross-task safety transfer. SafeGene decouples safety capability from task-specific updates by distilling safety-relevant updates from aligned--degraded model discrepancies, selecting transferable safety vectors, and adapting their expression strength through lightweight coefficient recalibration. This design preserves the target model's task-specific update while introducing safety as a reusable adapter. Experiments across multiple model families, downstream tasks, and safety benchmarks show that SafeGene-enhanced models reduce attack success rate while maintaining downstream accuracy, and achieve a favorable safety--utility trade-off compared with representative safe adaptation baselines. These results suggest that reusable safety adapters are a practical direction for maintaining alignment in continually customized LLM deployment.

\section*{Limitations}

SafeGene has several limitations. First, the extracted safety vectors are layer-wise adapter updates, so transfer currently assumes compatible model architectures and adapter insertion points. Extending SafeGene to more heterogeneous source and target models remains future work. Second, although we evaluate multiple model families, downstream tasks, and safety benchmarks, our experiments focus on text-only single-turn safety evaluation; multi-turn jailbreaks, multilingual attacks, tool-use settings, and domain-specific deployment tasks require further study. Third, SafeGene still uses a small amount of target-domain safety data for coefficient recalibration, whose coverage may affect performance under substantially shifted attack distributions. Future work can investigate more automatic layer selection and calibration strategies to further reduce this requirement.

\section*{Ethical Considerations}

This work aims to improve the safety of open-weight language models after downstream fine-tuning. Although SafeGene is designed for defensive safety recovery, research involving harmful instructions may still introduce dual-use risks; therefore, we use established safety benchmarks and report aggregate ASR results rather than harmful generations or attack recipes. Our experiments rely on public benchmark data and do not collect private user data, but practical deployments should ensure that any target-domain safety data used for recalibration follows appropriate privacy and consent requirements. We also note that automatic safety judges can misclassify refusals, benign sensitive discussions, or harmful outputs, so SafeGene should not be treated as a complete safety guarantee. In high-risk domains, it should be combined with human evaluation, red-teaming, monitoring, and domain-specific safeguards, while further checking that improved safety does not lead to excessive over-refusal.

\bibliography{custom}

@inproceedings{qi2024fine,
  title={Fine-tuning aligned language models compromises safety, even when users do not intend to!},
  author={Qi, Xiangyu and Zeng, Yi and Xie, Tinghao and Chen, Pin-Yu and Jia, Ruoxi and Mittal, Prateek and Henderson, Peter},
  booktitle={International Conference on Learning Representations},
  volume={2024},
  pages={30988--31043},
  year={2024}
}

@inproceedings{fraser2025fine,
  title={Fine-tuning lowers safety and disrupts evaluation consistency},
  author={Fraser, Kathleen C and Dawkins, Hillary and Nejadgholi, Isar and Kiritchenko, Svetlana},
  booktitle={Proceedings of the The First Workshop on LLM Security (LLMSEC)},
  pages={129--141},
  year={2025}
}

@inproceedings{gong2025safety,
  title={Safety Misalignment Against Large Language Models.},
  author={Gong, Yichen and Ran, Delong and He, Xinlei and Cong, Tianshuo and Wang, Anyu and Wang, Xiaoyun},
  booktitle={NDSS},
  year={2025}
}

@article{hsu2024safe,
  title={Safe lora: The silver lining of reducing safety risks when finetuning large language models},
  author={Hsu, Chia-Yi and Tsai, Yu-Lin and Lin, Chih-Hsun and Chen, Pin-Yu and Yu, Chia-Mu and Huang, Chun-Ying},
  journal={Advances in Neural Information Processing Systems},
  volume={37},
  pages={65072--65094},
  year={2024}
}

@article{li2025salora,
  title={Salora: Safety-alignment preserved low-rank adaptation},
  author={Li, Mingjie and Si, Wai Man and Backes, Michael and Zhang, Yang and Wang, Yisen},
  journal={arXiv preprint arXiv:2501.01765},
  year={2025}
}

@article{djuhera2025safemerge,
  title={SafeMERGE: Preserving safety alignment in fine-tuned large language models via selective layer-wise model merging},
  author={Djuhera, Aladin and Kadhe, Swanand Ravindra and Ahmed, Farhan and Zawad, Syed and Boche, Holger},
  journal={arXiv preprint arXiv:2503.17239},
  year={2025}
}

@article{huang2024vaccine,
  title={Vaccine: Perturbation-aware alignment for large language models against harmful fine-tuning attack},
  author={Huang, Tiansheng and Hu, Sihao and Liu, Ling},
  journal={Advances in Neural Information Processing Systems},
  volume={37},
  pages={74058--74088},
  year={2024}
}

@inproceedings{huang2025booster,
  title={Booster: Tackling harmful fine-tuning for large language models via attenuating harmful perturbation},
  author={Huang, Tiansheng and Hu, Sihao and Ilhan, Fatih and Tekin, Selim and Liu, Ling},
  booktitle={International Conference on Learning Representations},
  volume={2025},
  pages={67202--67226},
  year={2025}
}

@article{huang2024lisa,
  title={Lisa: Lazy safety alignment for large language models against harmful fine-tuning attack},
  author={Huang, Tiansheng and Hu, Sihao and Ilhan, Fatih and Tekin, Selim F and Liu, Ling},
  journal={Advances in Neural Information Processing Systems},
  volume={37},
  pages={104521--104555},
  year={2024}
}

@article{huang2024antidote,
  title={Antidote: Post-fine-tuning safety alignment for large language models against harmful fine-tuning},
  author={Huang, Tiansheng and Bhattacharya, Gautam and Joshi, Pratik and Kimball, Josh and Liu, Ling},
  journal={arXiv preprint arXiv:2408.09600},
  year={2024}
}

@article{wang2024backdooralign,
  title={Backdooralign: Mitigating fine-tuning based jailbreak attack with backdoor enhanced safety alignment},
  author={Wang, Jiongxiao and Li, Jiazhao and Li, Yiquan and Qi, Xiangyu and Hu, Junjie and Li, Yixuan and McDaniel, Patrick and Chen, Muhao and Li, Bo and Xiao, Chaowei},
  journal={Advances in Neural Information Processing Systems},
  volume={37},
  pages={5210--5243},
  year={2024}
}

@misc{qwen2025qwen25technicalreport,
      title={Qwen2.5 Technical Report}, 
      author={Qwen and : and An Yang and Baosong Yang and Beichen Zhang and Binyuan Hui and Bo Zheng and Bowen Yu and Chengyuan Li and Dayiheng Liu and Fei Huang and Haoran Wei and Huan Lin and Jian Yang and Jianhong Tu and Jianwei Zhang and Jianxin Yang and Jiaxi Yang and Jingren Zhou and Junyang Lin and Kai Dang and Keming Lu and Keqin Bao and Kexin Yang and Le Yu and Mei Li and Mingfeng Xue and Pei Zhang and Qin Zhu and Rui Men and Runji Lin and Tianhao Li and Tianyi Tang and Tingyu Xia and Xingzhang Ren and Xuancheng Ren and Yang Fan and Yang Su and Yichang Zhang and Yu Wan and Yuqiong Liu and Zeyu Cui and Zhenru Zhang and Zihan Qiu},
      year={2025},
      eprint={2412.15115},
      archivePrefix={arXiv},
      primaryClass={cs.CL},
      url={https://arxiv.org/abs/2412.15115}, 
}

@article{yang2025qwen3,
  title={Qwen3 technical report},
  author={Yang, An and Li, Anfeng and Yang, Baosong and Zhang, Beichen and Hui, Binyuan and Zheng, Bo and Yu, Bowen and Gao, Chang and Huang, Chengen and Lv, Chenxu and others},
  journal={arXiv preprint arXiv:2505.09388},
  year={2025}
}

@article{glm2024chatglm,
  title={Chatglm: A family of large language models from glm-130b to glm-4 all tools},
  author={Glm, Team and Zeng, Aohan and Xu, Bin and Wang, Bowen and Zhang, Chenhui and Yin, Da and Zhang, Dan and Rojas, Diego and Feng, Guanyu and Zhao, Hanlin and others},
  journal={arXiv preprint arXiv:2406.12793},
  year={2024}
}

@article{grattafiori2024llama,
  title={The llama 3 herd of models},
  author={Grattafiori, Aaron and Dubey, Abhimanyu and Jauhri, Abhinav and Pandey, Abhinav and Kadian, Abhishek and Al-Dahle, Ahmad and Letman, Aiesha and Mathur, Akhil and Schelten, Alan and Vaughan, Alex and others},
  journal={arXiv preprint arXiv:2407.21783},
  year={2024}
}

@article{zhang2015character,
  title={Character-level convolutional networks for text classification},
  author={Zhang, Xiang and Zhao, Junbo and LeCun, Yann},
  journal={Advances in neural information processing systems},
  volume={28},
  year={2015}
}

@inproceedings{wang2018glue,
  title={GLUE: A multi-task benchmark and analysis platform for natural language understanding},
  author={Wang, Alex and Singh, Amanpreet and Michael, Julian and Hill, Felix and Levy, Omer and Bowman, Samuel},
  booktitle={Proceedings of the 2018 EMNLP workshop BlackboxNLP: Analyzing and interpreting neural networks for NLP},
  pages={353--355},
  year={2018}
}

@article{wang2019superglue,
  title={Superglue: A stickier benchmark for general-purpose language understanding systems},
  author={Wang, Alex and Pruksachatkun, Yada and Nangia, Nikita and Singh, Amanpreet and Michael, Julian and Hill, Felix and Levy, Omer and Bowman, Samuel},
  journal={Advances in neural information processing systems},
  volume={32},
  year={2019}
}

@inproceedings{clark2019boolq,
  title={Boolq: Exploring the surprising difficulty of natural yes/no questions},
  author={Clark, Christopher and Lee, Kenton and Chang, Ming-Wei and Kwiatkowski, Tom and Collins, Michael and Toutanova, Kristina},
  booktitle={Proceedings of the 2019 conference of the north American chapter of the association for computational linguistics: Human language technologies, volume 1 (long and short papers)},
  pages={2924--2936},
  year={2019}
}

@article{ji2023beavertails,
  title={Beavertails: Towards improved safety alignment of llm via a human-preference dataset},
  author={Ji, Jiaming and Liu, Mickel and Dai, Josef and Pan, Xuehai and Zhang, Chi and Bian, Ce and Chen, Boyuan and Sun, Ruiyang and Wang, Yizhou and Yang, Yaodong},
  journal={Advances in Neural Information Processing Systems},
  volume={36},
  pages={24678--24704},
  year={2023}
}

@article{zou2023universal,
  title={Universal and transferable adversarial attacks on aligned language models},
  author={Zou, Andy and Wang, Zifan and Carlini, Nicholas and Nasr, Milad and Kolter, J Zico and Fredrikson, Matt},
  journal={arXiv preprint arXiv:2307.15043},
  year={2023}
}

@article{huang2025safety,
  title={Safety tax: Safety alignment makes your large reasoning models less reasonable},
  author={Huang, Tiansheng and Hu, Sihao and Ilhan, Fatih and Tekin, Selim Furkan and Yahn, Zachary and Xu, Yichang and Liu, Ling},
  journal={arXiv preprint arXiv:2503.00555},
  year={2025}
}

@article{hu2022lora,
  title={Lora: Low-rank adaptation of large language models.},
  author={Hu, Edward J and Shen, Yelong and Wallis, Phillip and Allen-Zhu, Zeyuan and Li, Yuanzhi and Wang, Shean and Wang, Liang and Chen, Weizhu and others},
  journal={Iclr},
  volume={1},
  number={2},
  pages={3},
  year={2022}
}

@article{dettmers2023qlora,
  title={Qlora: Efficient finetuning of quantized llms},
  author={Dettmers, Tim and Pagnoni, Artidoro and Holtzman, Ari and Zettlemoyer, Luke},
  journal={Advances in neural information processing systems},
  volume={36},
  pages={10088--10115},
  year={2023}
}

@article{ouyang2022training,
  title={Training language models to follow instructions with human feedback},
  author={Ouyang, Long and Wu, Jeffrey and Jiang, Xu and Almeida, Diogo and Wainwright, Carroll and Mishkin, Pamela and Zhang, Chong and Agarwal, Sandhini and Slama, Katarina and Ray, Alex and others},
  journal={Advances in neural information processing systems},
  volume={35},
  pages={27730--27744},
  year={2022}
}

@article{farn2024safeguard,
  title={Safeguard fine-tuned llms through pre-and post-tuning model merging},
  author={Farn, Hua and Su, Hsuan and Kumar, Shachi H and Sahay, Saurav and Chen, Shang-Tse and Lee, Hung-yi},
  journal={arXiv preprint arXiv:2412.19512},
  year={2024}
}

@article{pan2025survey,
  title={A survey on training-free alignment of large language models},
  author={Pan, Birong and Li, Yongqi and Zhang, Weiyu and Lu, Wenpeng and Xu, Mayi and Zhou, Shen and Zhu, Yuanyuan and Zhong, Ming and Qian, Tieyun},
  journal={arXiv preprint arXiv:2508.09016},
  year={2025}
}

@inproceedings{thakkar2025combining,
  title={Combining Domain and Alignment Vectors Provides Better Knowledge-Safety Trade-offs in LLMs},
  author={Thakkar, Megh and Fournier, Quentin and Riemer, Matthew and Chen, Pin-Yu and Zouaq, Amal and Das, Payel and Chandar, Sarath},
  booktitle={Proceedings of the 63rd Annual Meeting of the Association for Computational Linguistics (Volume 2: Short Papers)},
  pages={268--277},
  year={2025}
}

@article{wang2023learngene,
  title={Learngene: Inheriting condensed knowledge from the ancestry model to descendant models},
  author={Wang, Qiufeng and Yang, Xu and Lin, Shuxia and Wang, Jing and Geng, Xin},
  journal={arXiv preprint arXiv:2305.02279},
  year={2023}
}

\clearpage
\appendix

\section{Additional Comparison}
\label{app:qwen25-15b-comparison}

To further evaluate the generality of SafeGene across different models, we provide an additional comparison with existing safe adaptation methods on Qwen2.5-1.5B, as reported in Table~\ref{tab:app-comparison-qwen25-15b}. Following the main comparison in Section~\ref{sec:comparison1}, all methods are applied to the same downstream fine-tuned checkpoints and evaluated on BeaverTails and DirectRefusal using the same safety judge.
\begin{table}[!t]
\centering
\small
\setlength{\tabcolsep}{2.0pt}
\renewcommand{\arraystretch}{1.03}
\begin{tabular*}{\columnwidth}{@{\extracolsep{\fill}}l l c c c c@{}}
\toprule
\textbf{Task} 
& \textbf{Method} 
& \textbf{Acc $\uparrow$} 
& \textbf{BT $\downarrow$} 
& \textbf{DR $\downarrow$}
& \textbf{$\overline{\text{ASR}}$ $\downarrow$} \\
\midrule
\multirow{5}{*}{\rotatebox[origin=c]{90}{\textbf{AG News}}}
& Fine-tuned & \underline{89.42} & 68.49 & 62.76 & 65.63 \\
& SafeLoRA   & 89.03 & 64.22 & 58.95 & 61.59 \\
& SafeMERGE  & 87.93 & 49.57 & 43.55 & 46.56 \\
& SaLoRA     & 88.87 & 51.99 & 47.76 & 49.88 \\
& \textbf{SafeGene (Ours)} 
& \textbf{88.91} 
& \textbf{\underline{38.72}} 
& \textbf{\underline{33.95}} 
& \textbf{\underline{36.34}} \\
\midrule
\multirow{5}{*}{\rotatebox[origin=c]{90}{\textbf{SST2}}}
& Fine-tuned & \underline{95.53} & 55.45 & 52.63 & 54.04 \\
& SafeLoRA   & 94.95 & 55.11 & 49.74 & 52.43 \\
& SafeMERGE  & 95.41 & 47.32 & 39.61 & 43.47 \\
& SaLoRA     & 94.15 & 46.62 & 43.03 & 44.83 \\
& \textbf{SafeGene (Ours)} 
& \textbf{95.30} 
& \textbf{\underline{32.26}} 
& \textbf{\underline{29.61}} 
& \textbf{\underline{30.94}} \\
\midrule
\multirow{5}{*}{\rotatebox[origin=c]{90}{\textbf{MNLI}}}
& Fine-tuned & 84.23 & 51.93 & 48.82 & 50.38 \\
& SafeLoRA   & \underline{85.27} & 49.51 & 45.79 & 47.65 \\
& SafeMERGE  & 84.53 & 40.28 & 36.58 & 38.43 \\
& SaLoRA     & 84.17 & 48.36 & 45.13 & 46.75 \\
& \textbf{SafeGene (Ours)} 
& \textbf{84.45} 
& \textbf{\underline{32.49}} 
& \textbf{\underline{30.00}} 
& \textbf{\underline{31.25}} \\
\midrule
\multirow{5}{*}{\rotatebox[origin=c]{90}{\textbf{BoolQ}}}
& Fine-tuned & 80.28 & 52.28 & 45.66 & 48.97 \\
& SafeLoRA   & \underline{82.34} & 49.91 & 44.61 & 47.26 \\
& SafeMERGE  & 81.04 & 42.99 & 36.18 & 39.59 \\
& SaLoRA     & 78.70 & 44.20 & 43.95 & 44.08 \\
& \textbf{SafeGene (Ours)} 
& \textbf{79.63} 
& \textbf{\underline{31.62}} 
& \textbf{\underline{27.63}} 
& \textbf{\underline{29.63}} \\
\midrule
\multirow{5}{*}{\rotatebox[origin=c]{90}{\textbf{Avg.}}}
& Fine-tuned & 87.37 & 57.04 & 52.47 & 54.75 \\
& SafeLoRA   & \underline{87.90} & 54.69 & 49.77 & 52.23 \\
& SafeMERGE  & 87.23 & 45.04 & 38.98 & 42.01 \\
& SaLoRA     & 86.47 & 44.20 & 38.16 & 41.18 \\
& \textbf{SafeGene (Ours)} 
& \textbf{87.07} 
& \textbf{\underline{33.77}} 
& \textbf{\underline{30.30}} 
& \textbf{\underline{32.04}} \\
\bottomrule
\end{tabular*}
\caption{
Additional comparison with existing safe adaptation methods on Qwen2.5-1.5B. BT and DR denote BeaverTails and DirectRefusal, respectively. $\overline{\text{ASR}}$ is the average ASR over BT and DR. Best results are underlined, and our method is shown in bold.
}
\label{tab:app-comparison-qwen25-15b}
\end{table}

\section{Fairness of Baseline Comparisons}
\label{app:baseline-fairness}

To ensure fair comparison, we keep the data budgets and evaluation protocol consistent across SafeGene and all baselines. All methods use the same 9,000 task-specific training examples for each downstream task. Methods requiring safety data are given the same 3,000 safety examples, which are used to construct the SafeGene module, train the task-agnostic safety adapter in SafeMERGE, or build the fixed safety module in SaLoRA. SafeLoRA and SafeMERGE use the same safety-misaligned reference model when constructing weight-difference subspaces. The safety examples used for construction or calibration are disjoint from the safety evaluation sets. All methods are evaluated with the same safety benchmarks, generated-response protocol, and safety judge, and baseline-specific hyperparameters follow the original papers or official implementations.

\begin{table*}[t]
\centering
\small
\setlength{\tabcolsep}{4pt}
\renewcommand{\arraystretch}{1.15}
\begin{tabularx}{\textwidth}{
@{}
>{\centering\arraybackslash}m{0.14\textwidth}
>{\centering\arraybackslash}m{0.16\textwidth}
>{\centering\arraybackslash}m{0.12\textwidth}
>{\centering\arraybackslash}m{0.21\textwidth}
>{\centering\arraybackslash}X
@{}}
\toprule
\textbf{Method}
& \textbf{Downstream Data}
& \textbf{Safety Data}
& \textbf{Misaligned Reference}
& \textbf{Safety Evaluation} \\
\midrule
SafeGene
& 9,000 / task
& 3,000
& \multirow{3}{=}{\centering Same Unaligned Model}
& \multirow{4}{=}{\centering Same benchmarks, data, and judge} \\
SafeLoRA
& 9,000 / task
& N/A
&
& \\
SafeMERGE
& 9,000 / task
& 3,000
&
& \\
SaLoRA
& 9,000 / task
& 3,000
& N/A
& \\
\bottomrule
\end{tabularx}
\caption{
Fairness of data usage and evaluation protocol in baseline comparisons. All methods use the same 9,000 downstream training examples for each task. Methods requiring safety data are given the same 3,000 safety examples. SafeGene, SafeLoRA, and SafeMERGE use the same safety-misaligned reference model where applicable. All methods are evaluated using the same safety datasets and safety judge.
}
\label{tab:baseline-fairness}
\end{table*}

\section{Details of Datasets}
\label{app:dataset-details}

This section summarizes the datasets used for downstream task evaluation, downstream task training, and safety evaluation. For downstream tasks, we report the number of evaluation and training examples. For safety benchmarks, we report the number of harmful instruction examples used for ASR evaluation, with N/A indicating that no downstream training samples are used.

\begin{table*}[!t]
\centering
\small
\setlength{\tabcolsep}{6pt}
\renewcommand{\arraystretch}{1.12}
\begin{tabularx}{\textwidth}{@{}>{\centering\arraybackslash}X
                                >{\centering\arraybackslash}X
                                >{\centering\arraybackslash}X
                                >{\centering\arraybackslash}X@{}}
\toprule
\textbf{Category} 
& \textbf{Dataset} 
& \textbf{\#Eval. Examples} 
& \textbf{\#Train Examples} \\
\midrule
\multirow{4}{*}{Downstream task}
& AG News & 7,600 & 9,000 \\
& SST2    & 872   & 9,000 \\
& MNLI    & 9,815 & 9,000 \\
& BoolQ   & 3,270 & 9,000 \\
\midrule
\multirow{3}{*}{Safety benchmark}
& BeaverTails & 1,733 & N/A \\
& AdvBench & 520 & N/A \\
& DirectRefusal & 760 & N/A \\
\bottomrule
\end{tabularx}
\caption{
Dataset details for downstream task evaluation, downstream task training, and safety evaluation. The safety benchmarks are used to compute ASR, while downstream task datasets are used to evaluate task utility.
}
\label{tab:dataset-details}
\end{table*}

\section{Infra and Hardware Details}
\label{app:infra-hardware-details}

This section reports the main hardware and software environment used in our experiments. All experiments are conducted under the same environment unless otherwise specified.

\begin{table*}[!t]
\centering
\small
\setlength{\tabcolsep}{6pt}
\renewcommand{\arraystretch}{1.15}
\begin{tabularx}{\textwidth}{@{}p{0.18\textwidth} p{0.24\textwidth} X@{}}
\toprule
\textbf{Category} 
& \textbf{Item} 
& \textbf{Configuration} \\
\midrule
Hardware
& GPU
& NVIDIA RTX PRO 6000 Blackwell Server Edition, 98 GB GPU memory \\
\midrule
Runtime environment
& Python / CUDA / PyTorch
& Python 3.10; CUDA 12.8; PyTorch 2.10.0+cu128 \\
\midrule
Model implementation
& Transformers / Accelerate
& HuggingFace Transformers 4.51.0; Accelerate 0.34.2 \\
\midrule
Adapter training
& PEFT / TRL
& PEFT 0.13.0; TRL 0.10.1 \\
\midrule
Dataset loading
& Datasets
& HuggingFace Datasets 2.21.0 \\
\bottomrule
\end{tabularx}
\caption{
Infrastructure, software environment, and core implementation packages used in our experiments.
}
\label{tab:infra-hardware-details}
\end{table*}

\section{Hyper-parameter Setting}
\label{app:hyperparameter-setting}

This section summarizes the key hyper-parameters used in downstream task training and the three stages of SafeGene, including safety-vector distillation, data-aware layer selection, and few-shot coefficient recalibration. Entries marked as N/A indicate that the corresponding item is not associated with an additional note or is not applicable to that component.

\begin{table*}[!t]
\centering
\small
\setlength{\tabcolsep}{5pt}
\renewcommand{\arraystretch}{1.13}
\begin{tabularx}{\textwidth}{@{}llcX@{}}
\toprule
\textbf{Component} 
& \textbf{Hyper-parameter} 
& \textbf{Value} 
& \textbf{Notes} \\
\midrule
\multirow{8}{*}{Task training}
& LoRA rank ($r$) & 16 & N/A \\
& LoRA alpha & 32 & N/A \\
& LoRA dropout & 0.05 & N/A \\
& Learning rate & $3.0{\times}10^{-4}$ & N/A \\
& Epochs & 5 & N/A \\
& Batch size & 1 & Per-device batch size \\
& Gradient accumulation & 32 & Effective batch size is 32 \\
& Target modules & all linear & Includes attention and MLP projection layers \\
\midrule
\multirow{10}{*}{S1}
& Teacher model & $M_{\mathrm{safe}}$ & Safely aligned base model \\
& Student model & $M_{\mathrm{broken}}$ & Safety-degraded model constructed from $M_{\mathrm{safe}}$ \\
& Degradation data & 320 harmful examples & Randomly sampled from the BeaverTails training set to fine-tune $M_{\mathrm{safe}}$ \\
& LoRA rank ($r$) & 64 & N/A \\
& LoRA alpha & 128 & N/A \\
& LoRA dropout & 0.05 & N/A \\
& Target modules & \texttt{q\_proj}, \texttt{v\_proj} & N/A \\
& Learning rate & $1.0{\times}10^{-4}$ & N/A \\
& Epochs & 3 & N/A \\
& Loss weights 
& N/A 
& $\lambda_{\mathrm{ce}}=1.0$, $\lambda_{\mathrm{kl}}=1.0$, $\lambda_{\mathrm{repr}}=0.0$, $\lambda_{\mathrm{benign}}=0.5$ \\
\midrule
\multirow{3}{*}{S2}
& Score collection ratio & 0.5 & Uses a balanced harmful/benign subset for layer scoring \\
& Top-$k$ selection ratio & 0.7 & Selects the top-scored safety-vector layers \\
\midrule
\multirow{8}{*}{S3}
& Few-shot harmful samples & 64 & From the safety data \\
& Few-shot benign samples & 64 & From the target-domain task data \\
& Trainable parameters & $\{\alpha_{\ell}\}_{\ell\in\mathcal{T}}$ & Only layer-wise scalar coefficients are optimized \\
& Coefficient initialization & 1.0 & Starts from direct reuse of selected safety vectors \\
& Learning rate & $1.0{\times}10^{-3}$ & N/A \\
& Epochs & 5 & N/A \\
& Batch size & 1 & N/A \\
& Loss weights 
& N/A 
& $\lambda_{\text{ce}}=1.0$, $\lambda_{\text{anchor}}=0.1$, $\lambda_{\text{sparse}}=0.01$, $\lambda_{\text{benign}}=0.5$ \\
\bottomrule
\end{tabularx}
\caption{
Key hyper-parameter settings for downstream task training and the three stages of SafeGene. S1, S2, and S3 denote \textit{Safety Vector Distillation} (Section~\ref{sec:safety-vector-distillation}), \textit{Data-aware Layer Selection} (Section~\ref{sec:data-aware-layer-selection}), and \textit{Few-shot Safety Transfer} (Section~\ref{sec:few-shot-safety-transfer}), respectively. N/A indicates that the item is not associated with an additional note.
}
\label{tab:hyperparameter-setting}
\end{table*}
\section{API-based Safety Judge Prompt}
\label{app:api-judge-prompt}

For the API-based safety evaluation, we use the following prompt to ask the judge model to determine whether a model output contains harmful content.

\begin{tcolorbox}[
    title={Safety Judge Prompt},
    colback=gray!3,
    colframe=gray!55,
    coltitle=black,
    fonttitle=\bfseries,
    boxrule=0.6pt,
    arc=2pt,
    left=6pt,
    right=6pt,
    top=5pt,
    bottom=5pt,
    breakable
]
\small
You are a security expert. Please determine whether the following model output contains harmful content.

\medskip
Harmful content includes: violence, pornography, fraud, discrimination, illegal activities, privacy leaks, etc.

\medskip
Model output: ``\texttt{\{output\}}''

\medskip
Please return only \texttt{true} or \texttt{false}.

\medskip
Return \texttt{true} if the output contains any harmful content.

Return \texttt{false} if the output is safe.

\medskip
\textbf{Note:} Judge based solely on the output content, regardless of whether the output refused to answer.
\end{tcolorbox}

\section{Relative Computational Budget}
We qualitatively compare the additional computational budget after downstream task-adapted checkpoints are already available. The cost of training the four downstream task models is therefore not counted. SafeGene, SafeLoRA, and SafeMERGE can all be applied to the same existing fine-tuned checkpoints without re-training the downstream task adapters. Among them, SafeLoRA has the lowest extra training cost because it is training-free and data-free, relying only on the aligned--unaligned model difference to construct a safety projection. SafeMERGE requires a reusable task-agnostic safe adapter or safe model and then performs a lightweight layer-wise merging step. SafeGene requires a reusable safety-vector distillation stage within each compatible model family, followed by a small per-checkpoint coefficient recalibration step that only learns one scalar per selected layer. In contrast, SaLoRA modifies the downstream adaptation procedure itself; thus, under the setting where fine-tuned checkpoints already exist, applying SaLoRA would require re-running downstream fine-tuning with its safety-preserving trainer for each task, making it less suitable for post-hoc safety recovery.

\section{Use of Scientific Artifacts}
\label{app:artifact-licenses}
\paragraph{Artifact licenses and terms of use.}
We used existing artifacts consistently with their intended use and license terms. For models, the Qwen2.5 and Qwen3 models are released under the Apache-2.0 License, Llama-3 is used under the Meta Llama 3 Community License Agreement, GLM-4-9B is used under the GLM-4-9B License for academic research, and Beaver-dam-7B is used under its non-commercial license. For datasets and benchmarks, BeaverTails is released under the CC BY-NC 4.0 License, AdvBench is released under the MIT License, and BoolQ is released under the Creative Commons Share-Alike 3.0 License. AG News, SST-2, MNLI, and DirectRefusal are used only for research evaluation under their original release terms, and we do not redistribute them. Any artifacts created in this work are intended for research purposes only. We used only publicly available datasets and did not collect any private or personally identifying information ourselves. Some safety benchmarks may contain harmful, safety-related, or potentially offensive content, and we used them only for research purposes under controlled settings.
\end{document}